\documentclass[journal]{IEEEtran}
\usepackage{amsmath,amsfonts}
\usepackage{algorithmic}
\usepackage{algorithm}
\usepackage{array}
\usepackage[caption=false,font=normalsize,labelfont=sf,textfont=sf]{subfig}
\usepackage{textcomp}
\usepackage{stfloats}
\usepackage{url}
\usepackage{verbatim}
\usepackage{graphicx}
\usepackage{cite}
\usepackage{multirow}
\usepackage{arydshln}
\usepackage{mathrsfs}
\usepackage{bbm}
\usepackage{bm}

\begin{document}

\title{Understanding Convolutional Neural Networks \\ from Excitations}

\author{Zijian~Ying,
	Qianmu~Li,
	Zhichao~Lian,
	Jun~Hou,
	Tong~Lin,
	Tao~Wang

	\IEEEcompsocitemizethanks{
		\IEEEcompsocthanksitem Zijian~Ying, Zhichao~Lian are with the School of Cyber Science and Technology, Nanjing University of Science and Technology, Nanjing 210094, China. E-mail: zjying@njust.edu.cn, lzcts@163.com.
		\IEEEcompsocthanksitem Qianmu~Li is with the Digital Economy Research Institute, Nanjing University of Science and Technology, Nanjing 210094, China. E-mail: qianmu@njust.edu.cn.
		\IEEEcompsocthanksitem Jun~Hou is with Nanjing Vocational University of Industry Technology, Nanjing 210023, China. E-mail: houjunnjust@163.com.
		\IEEEcompsocthanksitem Tong~Lin, and Tao~Wang, are with the School of Computer Science and Engineering, Nanjing University of Science and Technology, Nanjing 210094, China. E-mail: 1466278404@qq.com, 122106010829@njust.edu.cn.}

	\thanks{Corresponding authors: Qianmu~Li and Zhichao~Lian.}}




\maketitle

\begin{abstract}
	Saliency maps have proven to be a highly efficacious approach for explicating the decisions of Convolutional Neural Networks. However, extant methodologies predominantly rely on gradients, which constrain their ability to explicate complex models. Furthermore, such approaches are not fully adept at leveraging negative gradient information to improve interpretive veracity. In this study, we present a novel concept, termed positive and negative excitation, which enables the direct extraction of positive and negative excitation for each layer, thus enabling complete layer-by-layer information utilization sans gradients. To organize these excitations into final saliency maps, we introduce a double-chain backpropagation procedure. A comprehensive experimental evaluation, encompassing both binary classification and multi-classification tasks, was conducted to gauge the effectiveness of our proposed method. Encouragingly, the results evince that our approach offers a significant improvement over the state-of-the-art methods in terms of salient pixel removal, minor pixel removal, and inconspicuous adversarial perturbation generation guidance. Additionally, we verify the correlation between positive and negative excitations.
\end{abstract}


\section{Introduction}
\begin{figure*}
	\centering
	\includegraphics[width=17.5cm]{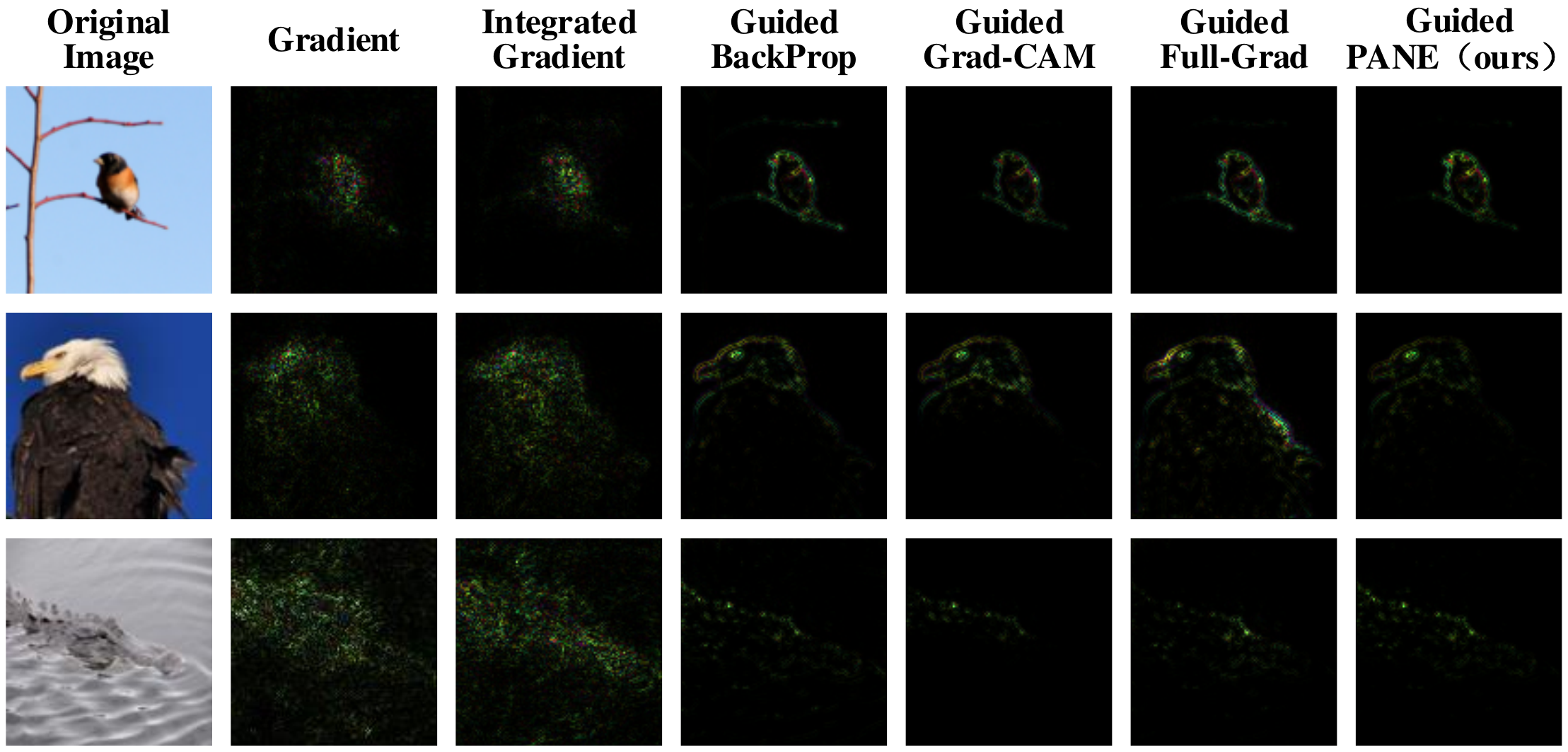}
	\caption{Visualization of saliency map from different methods. The saliency maps showcased in this figure stem from three inputs that were processed by a VGG-16 model trained on the ImageNet dataset. The Guided Grad-CAM, Guided Full-Grad, and Guided PANE are generated by the Grad-CAM, Full-Grad, and PANE methods were multiplied by the saliency maps obtained through Guided backpropagation, correspondingly. Compared to other methods, saliency maps from PANE focus more on discrete local features, i.e. certain pixels. From a statistical perspective, the saliency maps from PANE have a greater left-skewed distribution tendency.}
\end{figure*}

\IEEEPARstart{S}{aliency} map, pioneered by \cite{2014deepinsight}, has emerged as a widely used technique for explicating the decision-making processes of Convolutional Neural Networks. As a local interpretive method, saliency maps endeavor to identify the pixels or regions that exert the most significant influence on the model's decision for a given sample (i.e., image). However, some early methods, such as Guided backpropagation \cite{springenberg2014striving}, lack category sensitivity. One solution to this issue is to generate saliency maps based on perturbations. For instance, \cite{2014visualizing} utilizes fixed-size pixel blocks to mask the image serially, and gauges the saliency value of each block by observing changes in the target class. Conversely, \cite{fong2017interpretable} leverages optimization to facilitate mask learning, leading to meaningful perturbations. Nevertheless, perturbation-based saliency map generation methods suffer from considerable randomness, poor outcomes, and high computational complexity. To overcome these limitations and better characterize image saliency for a given category, Class Activation Mapping (CAM) \cite{2016learning} is proposed. CAM posits that the model's fully connected layers harbor high-dimensional semantic information (e.g., object concepts of categories) and that the feature maps generated by the convolutional layers contain spatial structure information. To make the saliency map category-sensitive, CAM obtains the mapping relationship from the target class to a specific layer of the feature map (i.e., linear weights). Subsequent CAM-based methods, abbreviated as CAMs, continue to primarily merge feature maps linearly to generate saliency (i.e., activation) maps.

However, a significant limitation of many existing CAMs, as well as other saliency map generation approaches, is their heavy reliance on gradients. Gradients are only able to produce accurate weights if the model is sufficiently linear, rendering gradient-based methods less effective when faced with more complex models. Moreover, the interpretation of negative gradients presents a significant challenge, prompting some researchers to use the ReLU function to filter out such gradients, as seen in \cite{springenberg2014striving} and \cite{2018gradcampp}. However, this approach results in further information loss within the network, which can negatively impact the accuracy of the saliency map.

In this study, we present a groundbreaking gradient-independent approach that addresses the aforementioned challenges. Our work introduces a novel concept to elucidate Convolutional Neural Networks, termed positive and negative excitation. By extracting the positive and negative excitations for each layer, we generate the positive and negative saliency maps via a double-chain backpropagation procedure. Either the positive saliency map alone or a combination of both contributes to producing a valid saliency map. The contributions of our work are summarized as follows:

\begin{itemize}
	\item We introduce a pioneering conceptualization known as positive and negative excitation (PANE). PANE extracts both positive and negative contributions for each layer to form the local interpretation. The derivations of multiple classical layers are showcased.

	\item We present a novel double-chain backpropagation procedure, which enables the computation of composite excitation coefficients between any two layers within the network. By leveraging this procedure, both positive and negative excitation maps can be obtained. This double-chain backpropagation procedure effectively reduces the time complexity associated with calculating complete excitations, bringing it to a comparable level as that of forward backpropagation.

	\item To gauge the effectiveness of positive excitation maps, commonly referred to as traditional saliency maps, we conduct a comprehensive comparative analysis across binary classification and multi-classification tasks. Our evaluation involves benchmarking against several state-of-the-art baselines, renowned for their performance in the field.

	\item We further investigate the interplay between the positive and negative excitation maps and their impact on saliency mapping. Through comparison experiments, visualizations, and downstream task guidance, we aim to gain insights into their collective impact.
\end{itemize}

The remainder of this paper is structured as follows. Section 2 provides an overview of the notations used in saliency map generation and the underlying motivation. In section 3, we introduce the PANE framework, including the positive and negative excitations in specific layers and the generation of positive and negative excitation maps using double-chain backpropagation. Section 4 delves into related work on saliency map generation methods. Section 5 presents the results of our experiments and corresponding analysis. Lastly, in section 6, we offer some concluding remarks and outline potential avenues for future research.

\section{Preliminary}
This section embarks by presenting the notation linked with the saliency map. Following this, the motivations for the proposed method are derived from two intuitive concepts prevalent in the existing saliency map methodologies.

\subsection{Problem Settings}
The convolutional neural network $F$ can generate a prediction result $Y$ for an image $X$. The saliency map aims to produce a map $M$ that is of the same size as the input image $X$ and indicates the contribution magnitude of each pixel in the image towards the prediction result. A precise definition is provided below. The input image is represented by $X \in R^{C \times W \times H}$, where $C$ signifies the number of channels, $W$ represents the image width, and $H$ represents the image height. Accordingly, the saliency map $M$ should also belong to $R^{C \times W \times H}$. However, some classical studies \cite{2017gradcam, 2016learning} disregard the channel dimension and concentrate solely on the image width and height. As a result, $M \in R^{W \times H}$ is also an acceptable representation. For a $K$-classification task, the prediction result is $Y = F(X)$, where $Y \in R^K$. The convolutional neural network $F$ comprises a total of $N$ layers, and we denote the $n$-th layer as $L_{n}$, where $n \in N$. Further, from the model execution perspective, we define the output of the $n$-th layer as $O_{n}$ and the parameters of the $n$-th layer as $\theta_{n}$. Specifically, the output of the $N$-th layer $O_{N}$ corresponds to $Y$. Thus, a saliency map approach involves utilizing $F$, $X$, and $O$ to determine $M$.

\subsection{Intuition ideas of existing saliency map methods}
Current saliency map acquisition methods primarily rely on gradients \cite{springenberg2014striving, 2017gradcam}. The rationale behind this approach is that gradient values can, to some extent, reflect the significance of each dimension of the sample points in the current feature space. However, this intuition is based on an assumption that the model operates close to a linear function. Consequently, when the model is not sufficiently linear, these methods' accuracy may be reduced. Here, we provide a simple case to elucidate this point. Consider the tuple $(t,k)$, where $z = t^2 + k^3$. At the point $(2,3)$, where $z = 31$, $\partial z / \partial t = 4$ and $\partial z / \partial k = 27$. It can be easy to reconstruct this equation as a linear expression $z = 2t + 9k$ at the point $(2,3)$. The coefficients of this linear expression, $2$ and $9$, signify the importance of $t$ and $k$ at this point. However, ratios of importance ($2/9$) and gradients ($4/27$) are not the same. This means that the gradients do not reflect the real importance. In reality, the gradients of a non-linear function, such as this one, do not always accurately represent the importance of the independent variable at most points.

Another prominent technique for generating saliency maps is to filter low-frequency information using the ReLU function. However, in these methods, low-frequency information is merely blurred into negative values, leading to potentially reduced saliency map accuracy. To illustrate this point, consider the function $z = t^2 + k^3$, where the gradient-based map generated at point $(-2, 1)$ is $(-4, 3)$. When the $(-4, 3)$ values pass through ReLU, the final map becomes $(0, 3)$, which implies that $t$ has almost no influence on $z$ at the point $(-2, 1)$. This outcome is evidently unreasonable.

\subsection{Motivation}

To address the aforementioned issues, a straightforward idea has emerged. This approach involves replacing the model linearly, or linearly-like, at the sample points while retaining negative values in the saliency map. By doing so, the original network $F$ can be reformulated as a linear equation $Y = w \cdot X_{vec} + b$, where $w \in R^{K \times (C \cdot W \cdot H)}$ denotes the linear coefficient, $X_{vec} \in R^{(C \cdot W \cdot H)}$ represents the vector form of $X$, and $b \in R^K$ denotes the bias. Naturally, the saliency map for output $Y_k$ and input $X$ can be directly obtained, i.e., $M_{X, Y_k} = Fold_X(w_{k})$, where the $Fold_*(\cdot)$ function transforms the input vector into the same shape as the target variable $*$.

However, two additional issues emerge. Firstly, linearizing the entire network directly is a challenging task, primarily due to the presence of nonlinear layers in the model and the bias of each layer. Nonlinear layers result in the output is not directly one-to-one with the input pixel points, while bias can impede the accuracy of the linear coefficient. Secondly, the actual significance of retaining negative values in the saliency map is unclear. Although negative values themselves represent the importance of the corresponding input component, the magnitude and sign properties of negative values are challenging to interpret compared to positive values. It is therefore inappropriate to interpret absolute values as significance and symbols as directions. Consequently, a simple yet effective method is necessary to resolve these two issues.

An intuitive approach to solve the first problem is to develop a gradient-independent method for generating local interpretations at each layer and subsequently combining these interpretations. Given the diverse properties exhibited by layer functions in deep learning, it becomes crucial to tailor the local interpretation method to suit each type of layer. To solve the second problem, the interpretation should have the ability to separate positive and negative impacts.

Then one more crucial problem is raised since the use of gradients and backpropagation is not applicable in this context.
This means there is not a natural 'mathematic chain' that can be used to transfer information from the output to the input. 
How to guide the connection between the local interpretations of each layer becomes challenging. 
If we directly compute the composition results for interpretations from each layer, the process becomes exceedingly complex. Therefore, a novel procedure that mimics the information transfer process of backpropagation while simultaneously reducing complexity is highly desirable.

\section{Methodology}

To tackle the linearization difficulty, as discussed in Section 2.3, we propose an approach that linearizes each layer $L_n$ of the network $F$ individually. Given that interpretation is localized, each layer's function merely needs to generate a linear substitution function at the sample point.
To address the unclear meaning of negative values, we further introduce the concept of positive and negative excitation (PANE). PANE decomposes each layer into three components: positive excitation, negative excitation, and zero excitation. These excitations are then sequentially linked together using chain transfer, similar to backpropagation, to construct the final saliency map.
This section initially presents the PANE concept and subsequently outlines the linearization and excitation extraction techniques for several commonly used layers. It then illustrates how the excitations of each layer are chained together to ultimately form the excitation for the entire network, i.e., the saliency map

\subsection{Positive and Negative Excitation}

We present a comprehensive definition of excitation, beginning with positive excitation. Positive excitation denotes the element that renders the symbolic output representation explicit. Negative excitation, on the other hand, weakens the output. A zero excitation denotes an element that has no impact on the result. Consider the equation $1 = 2 + (-1) + 0$, where $1$ is the output, possessing a positive sign and a value of 1. The equation comprises three elements: $2$, $-1$, and $0$. The element $0$ contributes nothing to the final output, either in terms of sign or value, and is a zero excitation. $2$ determines the symbolic positive sign of output $1$ and is a positive excitation. Conversely, $-1$ constrains the output to a larger positive value and is a negative excitation. Following this principle, an output $O$ can be decomposed into three components as follows:
\begin{equation}
	O = O^{+} + O^{-} + O^{0}
\end{equation}
Here, $O^+$ denotes the positive excitation of $O$, $O^-$ denotes the negative excitation of $O$, and $O^0$ signifies the zero excitation of $O$.

In the case of a linear process, the obtained output $o$ can be expressed as:
\begin{equation}
	o = w \cdot x
\end{equation}
where $o$ is a real number, $w \in R^V$ is the $V$-dimensional linear coefficient, and $x \in R^V$ is the $V$-dimensional vector input. We define the sign of $o$ as $Sign(o) \in {+,-,0}$. We can then rewrite the original inner product as:
\begin{equation}
	o = \sum_{i = 1}^{V} w_{i}\times x_{i} = \sum_{i = 1}^{V} Sign(w_{i}\times x_{i}) \times |w_{i}\times x_{i}|
\end{equation}
When $o \neq 0$, excitations can be obtained as follows:
\begin{equation}
	\begin{aligned}
		o^{+} = \{w_{i}\times x_{i} | Sign(w_{i}\times x_{i}) = Sign(o), i \in V\}, \\
		o^{-} =\{w_{i}\times x_{i} | Sign(w_{i}\times x_{i}) = -Sign(o), i \in V\}, \\
		o^{0} = \{w_{i}\times x_{i} | Sign(w_{i}\times x_{i}) = 0, i \in V\}.
	\end{aligned}
\end{equation}
When $o = 0$, those parts whose signs are non-zero and opposite cancel each other out. In this particular case, the positive and negative excitations are equivalent.

Moreover, we define any input $\{x_i|\{w_i \times x_i\} \in o^+\}$ as positive excitation signal $o^{+}_{x}$ and $\{x_i|\{w_i \times x_i\} \in o^-\}$ as the negative excitation signal $o^{-}_{x}$. We also define any coefficient $\{w_i|\{w_i \times x_i\} \in o^+\}$ as the positive excitation coefficient of $x_i$ and $\{w_i|\{w_i \times x_i\} \in o^-\}$ is defined as negative excitation coefficient of $x_i$. The positive and negative excitation coefficient sets are uniformly defined in the form of $Exc(o^{\{\cdot\}}_x)$, where $\{\cdot\} \in \{+,-\}$

\subsection{Excitations in specific layers}

To extract the PANE of each layer more effectively, it is essential to linearize each layer. In instance-level explanation, the input signal is fixed, resulting in a deterministic state for all signals in the entire network. Therefore, the input and output of any layer in the network are fixed as well. Consequently, when studying the influence of some parts of the input, all other parameters can be treated as known weights. In the following sections, we will demonstrate the linearization process and PANE derivation for some common layers in CNNs.

\subsubsection{Linear layer}
The linear layer, a crucial building block in machine learning, is typically employed as the fully connected layer in CNNs. We will begin by discussing this layer. The canonical expression of a linear layer is given by:
\begin{equation}
	Y = w \cdot X + b
\end{equation}
Here, $X$ denotes the input, $w$ represents the weight, $b$ is the bias, and $Y$ is the output. We assume that $X \in R^{V} $, $Y \in R^{W}$, $b \in R^{W}$, and $w \in R^{W \times V}$. For fixed $w$ and $b$, the portion that is genuinely correlated with $X$ in this equation is $Y' = Y-b$. As illustrated in the preceding section, the positive excitation signals of $Y'$ can be obtained. Moreover, each element of the positive excitation coefficient is defined as follows:
\begin{equation}
	\text{\textit{Exc}}(Y'^{+}_{X})_{i,j} = \left\{
	\begin{aligned}
		0, \,  if \,  \{ w_{i,j} \times X_{j} \} \notin {Y'^{+}}_{i} \\
		w_{i,j}, \,   if \,  \{ w_{i,j} \times X_{j} \} \in {Y'^{+}}_{i}
	\end{aligned}
	\right.
\end{equation}
Here, $i \in W$ and $j \in V$. We utilize $0$ to fill in the blank section of the positive excitation. Notably, $0$ serves as the zero element of the multiplicative group and the unit element of the additive group in the real number field. Therefore, it does not have an impact on the calculation of excitations. The negative excitation coefficient $\text{\textit{Exc}}(Y'^{-}_{X})$ can be determined in a similar manner. Given that any $Y$ has a fixed influence from bias $b$, the excitation of $Y'$ is, in essence, the excitation of $Y$.

\subsubsection{Convolution layer}
The convolutional layer is one of the most representative and critical layers in CNNs. The general expression of convolution is given by:
\begin{equation}
	Y = X * f
\end{equation}
Here, $X$ represents the input, $f$ denotes the convolution kernel, and $Y$ is the output. In image convolution, we can define $X \in R^{W \times H}$, $f \in R^{i \times j}$, and a stride parameter $s$, which indicates the sliding step size of the convolution kernel $f$. Then, the output $Y \in R^{([(W-i)/s]+1) \times ([(H-i)/s]+1)}$ (abbreviated as $Y \in R^{o_{1} \times o_{2}}$, where $o_{1} = [(W-i)/s]+1 $ and $o_{2} = [(H-i)/s]+1 $). This can be rewritten in linear form as follows:
\begin{equation}
	Y^{trans} = X^{trans} \cdot f^{trans}
\end{equation}
Here, $Y^{trans} \in R^{(o_{1} \cdot o_{2})}$, $X^{trans} \in R^{(o_{1} \cdot o_{2})\times(i \cdot j)}$, and $f^{trans} \in R^{i \cdot j}$. $Y^{trans}$ and $f^{trans}$ are the vector expansions of $Y$ and $f$, respectively. $X^{trans}$ is the complex expansion of $X$ according to $f$. The relationship between the elements of $X^{trans}$ and $X$ is expressed as follows:
\begin{equation}
	{X^{trans}}_{k,l} = X_{[k/o_{1}]\cdot s +[l/i], (k\%o_{2}) \cdot s + (l\%i)}
\end{equation}
Here, $k \in [0,i-1]$ and $l \in [0,j-1]$. Thus, the original convolution is equivalent to a linear layer. Subsequently, the excitation can be obtained in a similar manner to that of the linear layer. This form of vector expansion can often be observed in accelerated convolutional layer calculations \cite{li2021involution}. The input that is expanded into a vector can be restored to its original shape by directly using deconvolution \cite{long2015fully}.

We also derived a tensor form of excitation. For any element $Y_{k,l} \in Y$, where $k \in [0, o_{1}-1]$ and $l \in [0, o_{2}-1]$, it is calculated only by a certain slice of $X$. The process is as follows:
\begin{equation}
	Y_{k,l} = \sum_{i = 0}^{m-1}\sum_{j = 0}^{n-1} X_{k \cdot s + i, l \cdot s + j} \cdot f_{i,j}
\end{equation}
The size of the slice is equal to the size of $f$. Thus, the excitation of each element in output $Y$ is the corresponding slice from input $X$. The specific excitation coefficient can be obtained as a linear function. When we explicitly embody the spatial information in the excitation, the corresponding excitation can be expressed as $ \text{\textit{Exc}}(Y^{{\cdot}}{X}) \in R^{o{1} \times o_{2} \times W \times H}$, where the value of the same place as the slice position is equal, and the other places are $0$.

\begin{figure*}
	\centering
	\includegraphics[width=17.8cm]{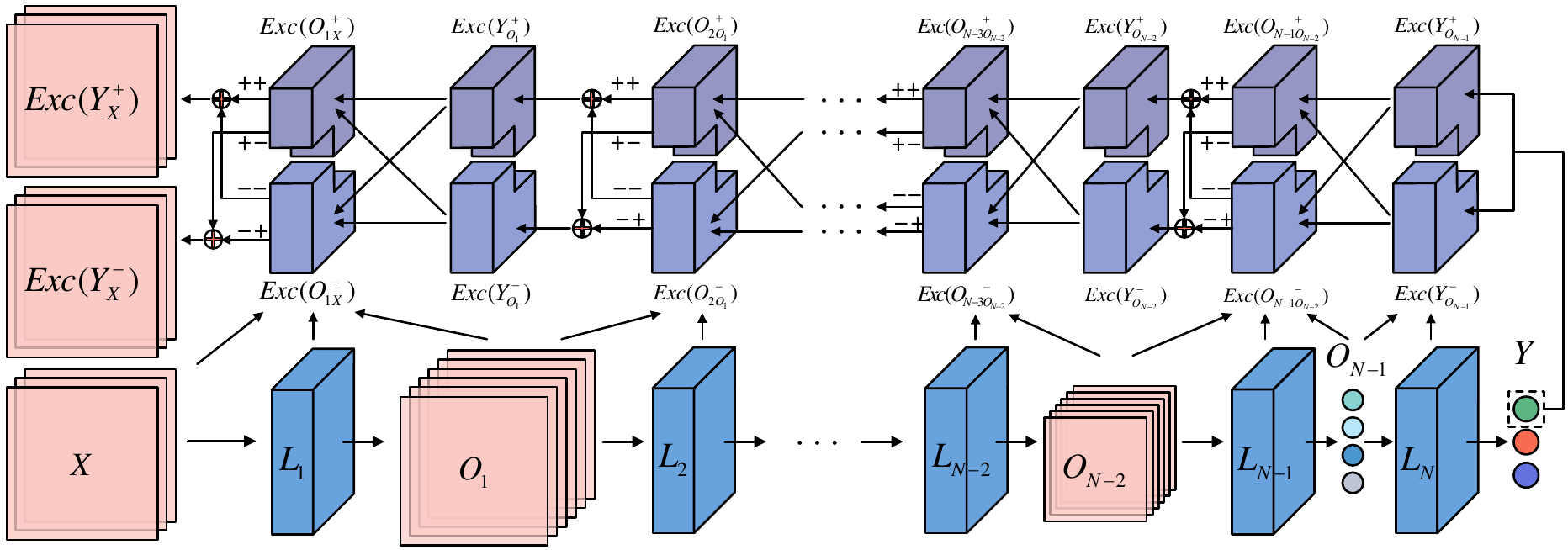}
	\caption{The process of excitation extraction. Following the input of samples into the network, the excitation coefficients for each layer's input to the output can be obtained from the input of that layer, the function of that layer, and the output of that layer, i.e., the blocks directly above each layer. The block directly above the output represents the excitation coefficient of the corresponding output to $Y$. In the blocks separated above, the lower block denotes the negative excitation coefficient, and the upper block represents the positive excitation coefficient. The block above the output is formed by backpropagation through a double-chain pass. The symbol '++' signifies that the excitation coefficient is formed by the compound of two positive excitation coefficients. Similarly, '+-', '--', and '-+' denote the combination of positive and negative excitation coefficients. $\oplus$ represents the summation of the two excitation coefficients.}\label{fig:attention}
\end{figure*}

\subsubsection{Activation layer}
Activation functions serve to filter input signals and rectify them into predetermined output distributions. Typically, activation functions are nonlinear and require the most linear function substitution. At the instance level, the input and output are fixed, and the function mapping process can be viewed as a linear coefficient. When the activation function operates on only one element of the input at a time and does not introduce any external influence, the coefficient is simply the output result divided by the input.

In the context of convolutional neural networks (CNNs), the rectified linear unit (ReLU) is a commonly used activation function. The ReLU function is defined as:
\begin{equation}
	ReLU(x) = max(x,0)
\end{equation}
where $x \in \mathbb{R}$. The function is linear when $x > 0$ and masks negative values in the input signal. The linear coefficient of $ReLU(x)$ is given by $\max(x,0)/x$, where $x\neq0$. It is noteworthy that ReLU has no negative excitation, and signals with values greater than zero are considered positive excitation signals. All positive excitation signals have a coefficient of 1, while the negative excitation coefficient of ReLU is set to zero for later calculations in this specific case. Nevertheless, this approach still adheres to the principle outlined in Equation (6).

\subsubsection{Pooling layer}
The pooling layer is a crucial component in expanding the receptive field of an input signal. In this regard, the two most common pooling methods are Max Pooling and Average Pooling. Interestingly, the pooling layer and the convolution layer are equivalent, and this duality is worth exploring.

Max Pooling, for instance, is defined as:
\begin{equation}
	MaxPool(X) = max(X)
\end{equation}
where $X \in \mathbb{R}^{W \times H}$ and $MaxPool(\cdot) \in \mathbb{R}$. This method selects the maximum value of a given area to represent the original entire input signal $X$. Similar to ReLU, Max Pooling only has positive excitation and zero excitation. The only positive excitation signal of $MaxPool(X)$ corresponds to the element $X_{i,j} \in X$ where $X_{i,j} = \max(X)$, and the positive excitation coefficient for this element is 1.

On the other hand, Average Pooling behaves more like a convolutional layer, where all values of the convolutional kernel are the same. The expression for Average Pooling is given by:
\begin{equation}
	AveragePool(X) = \bar{X} = \dfrac{\sum_{i=1,j=1}^{W,H}X_{i,j}}{W \times H}
\end{equation}
This is analogous to a linear function with all weights set to $(W \times H)^{-1}$ and the bias set to $\textbf{0}$. The excitation can be computed using the same principle as before.

\subsubsection{Normalization layer}
Normalization is a key technique that involves shifting the data distribution. For a fixed input signal $X$, the resulting output $Norm(X)$ has a mapping between each corresponding element in $X$ and $Norm(X)$, such that $Norm(X)$ can build a linear mapping to directly obtain $X$. In this context, we discuss the commonly used Batch Normalization (BN) technique as an example. The expression for BN is given by:
\begin{equation}
	BN(X) = \dfrac{X-\mu}{\sqrt{\sigma} + \epsilon} \circ \gamma + \beta
\end{equation}
where $X \in \mathbb{R}^{N}$, $\mu \in \mathbb{R}$ is the mean value of $X$, $\sigma \in \mathbb{R}$ is the variance of $X$, $\epsilon$ is a small bias to avoid division by zero, $\circ$ denotes the Hadamard product, $\gamma \in \mathbb{R}^{N}$ is a learned weight, and $\beta \in \mathbb{R}^{N}$ is the bias for output. For a given $X$, $\mu$ and $\sigma$ are fixed, and can therefore be treated as fixed parameters in this static analysis. The expression can then be rewritten as a linear function. For any element in the output $BN(X)$, we have:
\begin{equation}
	BN(X)_{i} = \dfrac{\gamma_{i}}{\sqrt{\sigma}+\epsilon} \cdot X_{i} + \beta_{i} - \dfrac{\mu \gamma_{i}}{\sqrt{\sigma}+\epsilon}= w'_{i} \cdot X_{i} + b'_{i}
\end{equation}
where $w'{i} = \dfrac{\gamma{i}}{\sqrt{\sigma}+\epsilon}$ and $b'{i} = \beta{i} - \dfrac{\mu \gamma_{i}}{\sqrt{\sigma}+\epsilon}$. The BN layer can obtain the excitation coefficients in the same way as a linear layer.

\subsection{Excitation for the entire net}

In any network $F$, the inputs and outputs of a given layer $L_n$ are denoted by $O_{n-1}$ and $O_{n}$, respectively, where $O_{n-1}$ is equal to $X$ when $n = 1$. Consequently, we can obtain the positive excitation coefficient $Exc({O_n}_{O{n-1}}^{+})$ and negative excitation coefficient $Exc({O_n}_{O{n-1}}^{-})$. Since zero excitation does not contribute to the output, it is not considered in subsequent analysis. Therefore, for all $N$ layers in the network, a total of $2N$ excitation coefficients can be obtained. It is worth noting that the positive excitation signal of positive excitation is positive, while the positive excitation signal of negative excitation is negative. This property of excitation signals is transitive and can be expressed as:
\begin{equation}
	\begin{aligned}
		{O_n}_{O_{n-2}}^{+} = {O_n}_{O_{n-1}}^{+} \cdot {O_{n-1}}_{O_{n-2}}^{+} \\
		+ {O_n}_{O_{n-1}}^{-} \cdot {O_{n-1}}_{O_{n-2}}^{-}
		\\
		{O_n}_{O_{n-2}}^{+} = {O_n}_{O_{n-1}}^{+} \cdot {O_{n-1}}_{O_{n-2}}^{-} \\
		+ {O_n}_{O_{n-1}}^{-} \cdot {O_{n-1}}_{O_{n-2}}^{+}
	\end{aligned}
\end{equation}
This transitivity is embodied in a double-chain structure, where the cross-multiplication of the respective positive and negative excitation signals of the two layers is followed by summation, thereby completing the transfer of the excitations. Since excitation coefficients have the same transitivity as excitation signals, they can also be transferred in this manner. Thus, the positive and negative excitation coefficients from any layer output $O_j$ to any layer output $O_i$ can be obtained iteratively, as follows:
\begin{equation}
	\begin{aligned}
		Exc({O_i}_{O_{j}}^{+}) = Exc({O_i}_{O_{i-1}}^{+}) \cdot  Exc({O_{i-1}}_{O_{j}}^{+}) \\
		+ Exc({O_i}_{O_{i-1}}^{-}) \cdot  Exc({O_{i-1}}_{O_{j}}^{-})                        \\
		Exc({O_i}_{O_{j}}^{-}) = Exc({O_i}_{O_{i-1}}^{+}) \cdot  Exc({O_{i-1}}_{O_{j}}^{-}) \\
		+ Exc({O_i}_{O_{i-1}}^{-}) \cdot  Exc({O_{i-1}}_{O_{j}}^{+})                        \\
	\end{aligned}
\end{equation}
where, $i > j$. When $i=j + 1$, $Exc({O_i}_{O_{j}}^{+}) = Exc({O_i}_{O_{i-1}}^{+})$ and $Exc({O_i}_{O_{j}}^{-}) = Exc({O_i}_{O_{i-1}}^{-})$.
When $i = N$ and $j = 0$, the resulting excitation coefficient is  $Exc({O_i}_{O_{j}}^{\{\cdot\}}) = Exc({Y}_{X}^{\{\cdot\}})$, representing the excitation coefficient for the entire network. $Exc({Y}_{X}^{\{\cdot\}})$ is the saliency map that we aim to obtain, where $Exc({Y}_{X}^{+})$ represents the coefficient that has a positive contribution to the result for the corresponding pixel in the image, while $Exc({Y}_{X}^{-})$ represents the opposite. Figure 2 illustrates the process involved in extracting the excitation coefficients for the entire network.

\subsection{Time complexity}

The computational efficiency of PANE is determined by two primary components: the extraction of excitation coefficients for each layer and the double-chain transfer of these coefficients. Notably, extracting the excitation coefficients is akin to coding layer functions, which is similar to constructing a network. As such, this process does not entail a significant computational time overhead.

On the other hand, the double-chain transfer of the excitation coefficients is analogous to the forward propagation of the network. The time complexity of network forwarding is represented by T($m$), and the time complexity of coefficient propagation is 4 times T($m$) due to the presence of double chains. Additionally, since there are two additive processes in the coefficient propagation process after each layer, the actual complexity is 4 times T($m$) + 2 times O($N$). It is worth noting that T($m$) is usually much larger than O($N$). Consequently, the time complexity can be approximated as 4 times T($m$).

\section{Related Work}
\subsection{Perturbation-based Saliency Map}
The generation of saliency maps is among the most effective approaches for providing local explanations of Convolutional Neural Networks. This technique involves three main directions, with the most straightforward approach entailing the use of perturbations to produce saliency maps. The core concept of these methods is to assess the saliency of a pixel or region by adding perturbations and observing the resulting changes in the network's predictions. One of the most classic works in this area is \cite{2014visualizing}, which employs a fixed-size block of pixels to sequentially shade each image from top to bottom and left to right while monitoring the changes in the predicted results. The more significant the perturbation's effect on the prediction result of the targeted class, the more salient the pixel at that location is deemed to be by the model.

While \cite{zhou2014object} considers random-value perturbations to be more reasonable than fixed-value pixel blocks, \cite{fong2017interpretable} asserts that masks, i.e., perturbations, are learnable and should not simply be panned or randomized. This work leverages optimization to clarify the actual meaning of the mask and perturb the pixels more effectively. Other approaches utilize generative models to perturb the image and generate saliency maps. For instance, \cite{agarwal2019removing} and \cite{chang2018explaining} create more visually appealing saliency maps by perturbing the features and using generative model repair. Meanwhile, \cite{fong2019understanding} adopts an extremal perturbations strategy to generate a saliency map with tighter bounds, and \cite{wagner2019interpretable} employs an adversarial attack approach to generate saliency graphs with finer granularity.

Despite their effectiveness, perturbation-based methods generally suffer from poor results and high time complexity.

\subsection{Gradient-based Saliency Map}
The gradient-based saliency map generation method leverages the gradient information in model backpropagation as a reference for pixel saliency. The most direct approach is to use the input gradient as the saliency map, which is also called Vanilla backpropagation (VBP) \cite{baehrens2010explain,2014deepinsight}. Guided backpropagation \cite{springenberg2014striving} applies the ReLU function to filter out negative gradients that are challenging to interpret. However, relying solely on gradients may produce excessive noise points on the saliency map. Moreover, the gradient conveys sensitivity rather than importance. \cite{smilkov2017smoothgrad} argues that the noise in the gradient is due to the unsmoothed objective function in the model learning process, and thus, these noises have no practical meaning. To address this issue, \cite{smilkov2017smoothgrad} proposed a method called Smooth Grad, which smooths the gradient's noise points to yield better visualization.

Integrated Gradient \cite{IntergratedGradients} is another method that corrects the gradient values in backpropagation. \cite{kim2019saliency} posits that noise arises from the propagation itself and suppresses non-significant features in the salient graph by setting a threshold value. Full-Grad \cite{2019fullgrad} substitutes the various gradient terms of the neural network output to generate the saliency map. \cite{9316988} proposes a differentiated relevance estimator, which introduces skewed distribution to estimate relevance scores.

LRP \cite{bach2015pixel} constructs the correlation between each layer and redefines the backpropagation method. Although LRP and the proposed method may appear similar, they are fundamentally different. Firstly, LRP still employs gradients to model correlations, while the proposed approach is gradient-independent. Secondly, LRP's backpropagation is layer-dependent, meaning that the backpropagation result of the previous layer must be computed before continuing with the current layer's dependency. In contrast, the proposed method is independent of each layer, which is more faithful to each layer's function. Finally, the positive and negative excitations in the proposed method are independent, whereas, in LRP, the positive and negative contributions are merely representations of the values in a contribution matrix. Importantly, LRP is based on the assumption of contribution conservation, while the proposed approach linearizes each layer in the local interpretation setting. Therefore, LRP and the proposed approach belong to completely different fields.

\subsection{Class Activation Map}
Class Activation Maps (CAMs) have emerged as one of the most widely used methods for generating salient maps on CNNs in recent years. The CAM approach leverages region-level features to highlight the regions most relevant to a particular category. This method is based on the assumption that fully connected layers contain more semantic information, while convolutional layers contain more spatial information \cite{2016learning}. Additionally, the closer to the output, the richer the semantic information, and the closer to the input, the more detailed the structural information. Therefore, the CAM approach typically involves obtaining a linear weighted aggregation of the feature maps closest to the fully connected layer to generate the salient map. \cite{2016learning} proposes global average pooling (GAP) to obtain these linear weights. However, since most mainstream models do not have the structure of GAP, Grad-CAM \cite{2017gradcam} uses gradient information to derive these linear weights. Specifically, Grad-CAM calculates the gradients of the specified feature maps by backpropagation and performs a simple aggregation of these gradients to generate linear weights between the feature maps. Grad-CAM++ \cite{2018gradcampp} improves on Grad-CAM by leveraging higher-order derivatives. Grad-CAM++ further enhances visualization by combining the weights of the channels.
Score-CAM \cite{2020scorecam} scores the forward propagation of certain classifications to derive the weights of the feature maps. Notably, Score-CAM is the first gradient-independent CAM-based method. Similarly, Ablation-CAM \cite{2020ablationcam} employs ablation analysis to determine the saliency of each pixel on the feature map, effectively realizing a "gradient-free" approach. LIFT-CAM \cite{jung2021towards} reconstructs the linear representation between feature maps based on DeepLIFT \cite{shrikumar2017learning} to further optimize the saliency map.
Other recent works on CAM include \cite{2020axiom, belharbi2022f, 2021layercam, lee2021relevance}.

\subsection{Other visualization explanations}
To gain insight into the network, some works try to inverse CNNs and generate visualization from the decision. One most recent works is \cite{9457245}, which proposes an inverse-based method to infer the corresponding region for the decision of the networks. This method inverses signals from different activations individually and highlights the salient regions with these signals. When it comes to the image itself, salient region, which also refers to segment to a certain extent, plays an important role in salient object detection (SOD). One of the most recent works is \cite{9560713}, which utilizes color channels and establishes multicolor contrast extraction mechanisms. It proposes a Siamese densely cooperative fusion (DCF) network, containing both boundary-directed feature learning and DCF, to detect saliency. \cite{9940193}, another latest work in SOD, observes that most existing saliency map has blurry regions. Then it tries to improve the clarity of the saliency region by designing a pixel value loss.

\section{Experiment}

In this study, we conducted a comprehensive set of experiments to investigate the correlation between saliency maps and model predictions, also known as faithfulness. To this end, we performed experiments on both binary-classification and multi-classification tasks.
For the binary classification task, we utilized the Dogs. vs. Cats dataset \cite{kaggle_dogs_vs_cats_dataset}, which comprises 25,000 images of cats and dogs. Dogs. vs. Cats dataset, one of the typical binary classification task datasets, comes from a competition on Kaggle. Images in this dataset have different shapes. Thus, when networks reshape images into the same size, the objects in the image will deform to a certain extent. Then, these dogs and cats with certain similar appearances contain different spatial and semantic information, which is conducive to revealing the relationship between classes. We randomly selected 1000 images from this dataset for the experiment.
For the multi-classification task, we employed the ImageNet(ILSVRC) \cite{russakovsky2015imagenet} dataset, which contains one image for each of the 1000 classes, totaling 1000 images. ImageNet \cite{deng2009imagenet}, which contains over 14 million images, is one of the most representative multi-classification task datasets. ImageNet(ILSVRC), a light version of ImageNet, comes from a competition named ImageNet Large Scale Visual Recognition Challenge in 2012. This dataset is currently widely used to evaluate the model structure and other works.
We used VGG-16 and AlexNet models that were pre-trained on the ImageNet dataset and fine-tuned them to maintain 100\% accuracy on the corresponding dataset.

To compare the performance of different saliency map generation methods, we selected Grad-CAM \cite{2017gradcam}, Grad-CAM++ \cite{2018gradcampp}, Score-CAM \cite{2020scorecam}, Full-Grad \cite{2019fullgrad}, and Guided Backpropagation (Guided BP)\cite{springenberg2014striving} as baselines. Grad-CAM and Guided BP are classic models, while Grad-CAM++ is the most widely used improved version of Grad-CAM. Score-CAM is representative of saliency map generation methods that do not use gradients, while Full-Grad is a method that fully utilizes gradients and can better represent the heat map generated by the gradient-based method.

To ensure that sufficient spatial structure information and high-dimensional semantic information were guaranteed, we set the target layer of all CAMs methods to be the last convolutional layer \cite{2017gradcam}. Since our proposed method, namely PANE, generates both positive and negative excitation activation maps, we solely used the positive activation map for comparison with the baselines in our experiments. Furthermore, we experimentally verified the positive and negative excitation activation maps and their interactions separately.

\begin{figure}
	\centering
	\includegraphics[width=8.7cm]{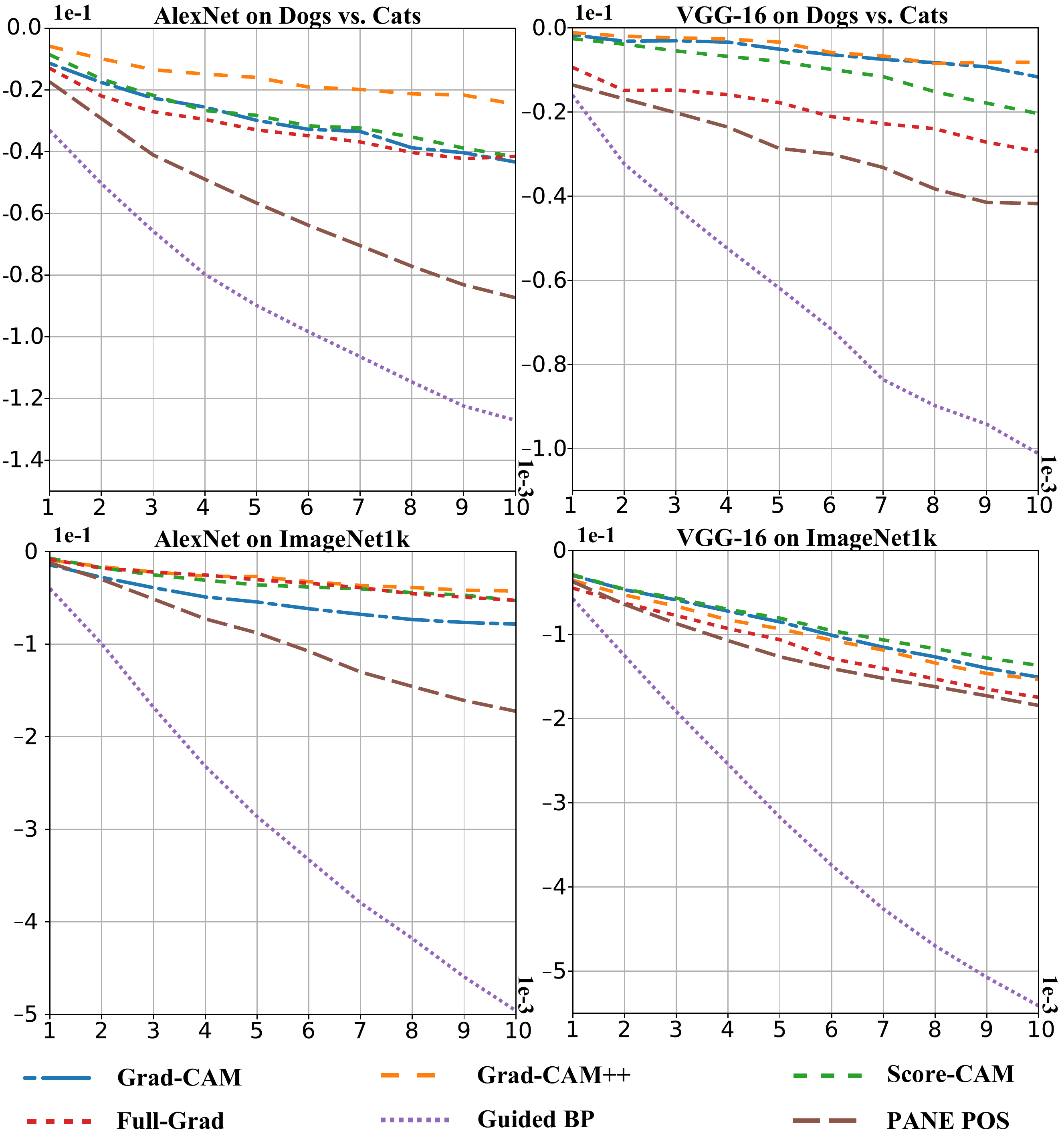}
	\caption{Experimental results on removing salient pixels. The network and dataset corresponding to the results of that experiment are indicated above each subplot.}
\end{figure}

\subsection{Remove salient pixels}

When a saliency map is faithful, it accurately reflects the contribution of individual pixels in the original image to the model prediction results. Consequently, removing the pixels marked as significant in the saliency map should result in a drop in the model's output. This experiment has been conducted in many previous studies \cite{2019fullgrad, 2021layercam}. However, saliency maps provide only a local interpretation and reflect the model's interpretation of the image at a particular sample point. As a result, removing a large number of pixels can cause the perturbed image to be too far from the original image in feature space. To address this issue, we limited the number of deleted pixels to less than 1\%. To evaluate the model's predicted changes after the pixels were removed, we used average probability drop (APD) as the evaluation metric, which is consistent with many existing works.

APD can effectively assess the changes in the model's prediction probabilities after pixels are removed from the image. As salient pixels are removed, the model's prediction probability drops. The greater the number of salient pixels removed, the more significant the prediction probability drop will be. Figure 3 displays the experimental results.

The results show that Guided BP seems to outperform all other methods. Except for the Buided BP, PANE-POS outperforms all other baselines no matter the ratios of 0.1\% or 1\%. Jointly observing all four subfigures, with the greater region, the gap between methods becomes more obvious. Overall, Guided BP performs best on most intervals, with PANE-POS coming in a close second. Compared to other methods, PANE-POS still maintains a large degree of leadership.

\begin{figure}
	\centering
	\includegraphics[width=8.7cm]{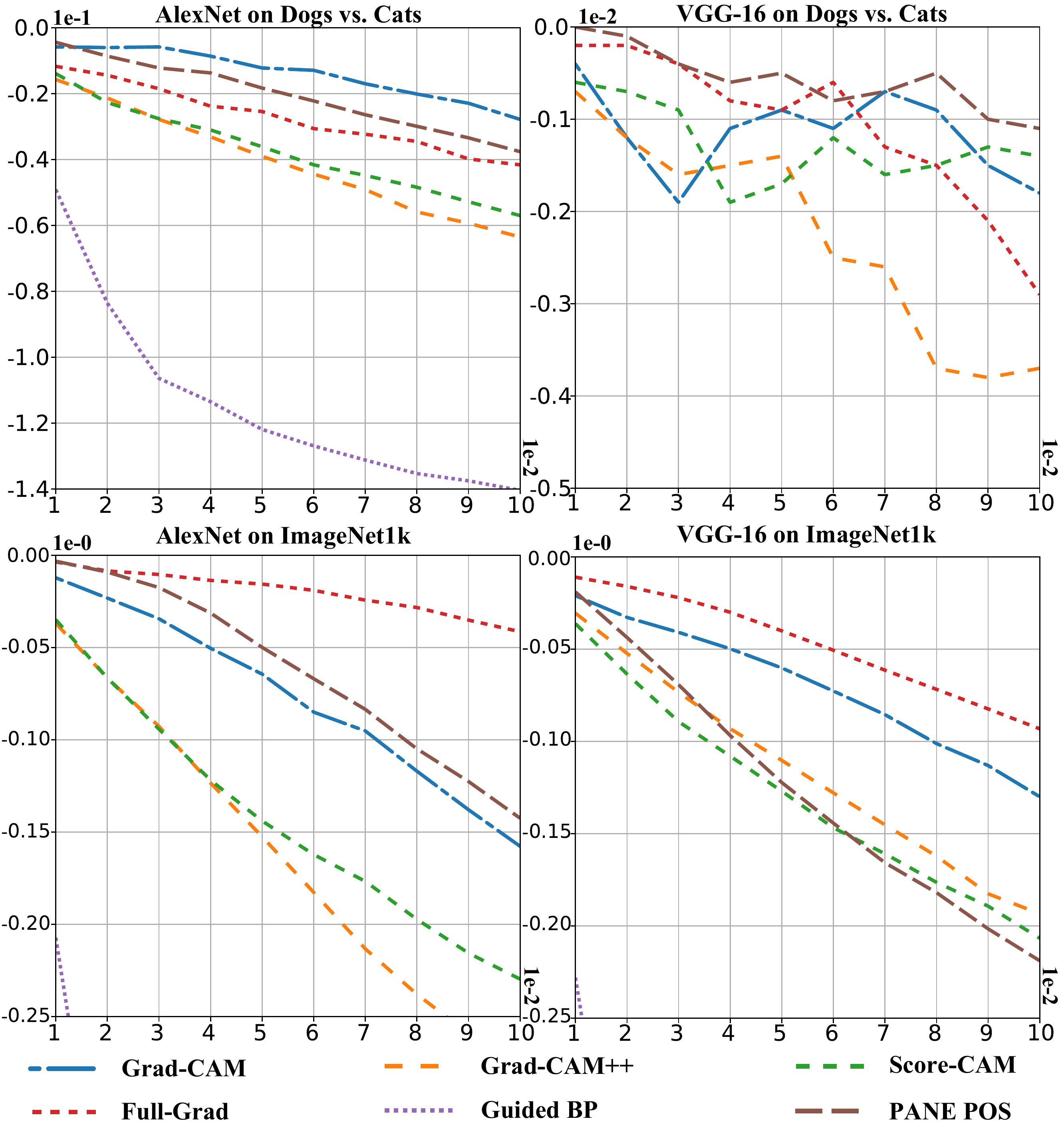}
	\caption{Experimental results on removing minor pixels. The network and dataset corresponding to the results of that experiment are indicated above each subplot.}
\end{figure}

\subsection{Remove minor pixels}

In their study, \cite{2019fullgrad} observed that replacing image pixels with black pixels can result in high-frequency edge artifacts that may cause output variation. To address this issue, they proposed a strategy to remove unimportant pixels to assess the performance of the saliency map. While we believe that very low perturbation areas can avoid the drawbacks caused by high-frequency edge artifacts, we still utilized the strategy proposed in \cite{2019fullgrad} for comparison. We used average probability drop as the evaluation metric, where smaller values indicate superior performance. We limited the number of deleted pixels to less than 10\%. It is worth to be noticed that the minor pixels in PANE are those excitation coefficients close to $0$. The experimental results are presented in Figure 4.

The best-performed method in the previous experiment, which is Guided BP, has an extremely poor performance. The APDs of Guided BP in each subfigure are all at the bottom. Because many of APDs are too great, the line even exceeds the lower bounds of the subfigure. PANE-POS still has good performance on all four subfigures. Although Grad-CAM and Full-Grad have advantages in some conditions, PANE-POS outperforms all other methods in the low perturbation interval.

Jointly considering the experimental results in Fig. 3 and Fig. 4, it can be inferred that PANE-POS exhibits superior performance in both evaluation metrics.

\subsection{Internal explanation}

\begin{figure*}
	\centering
	\includegraphics[width=17.5cm]{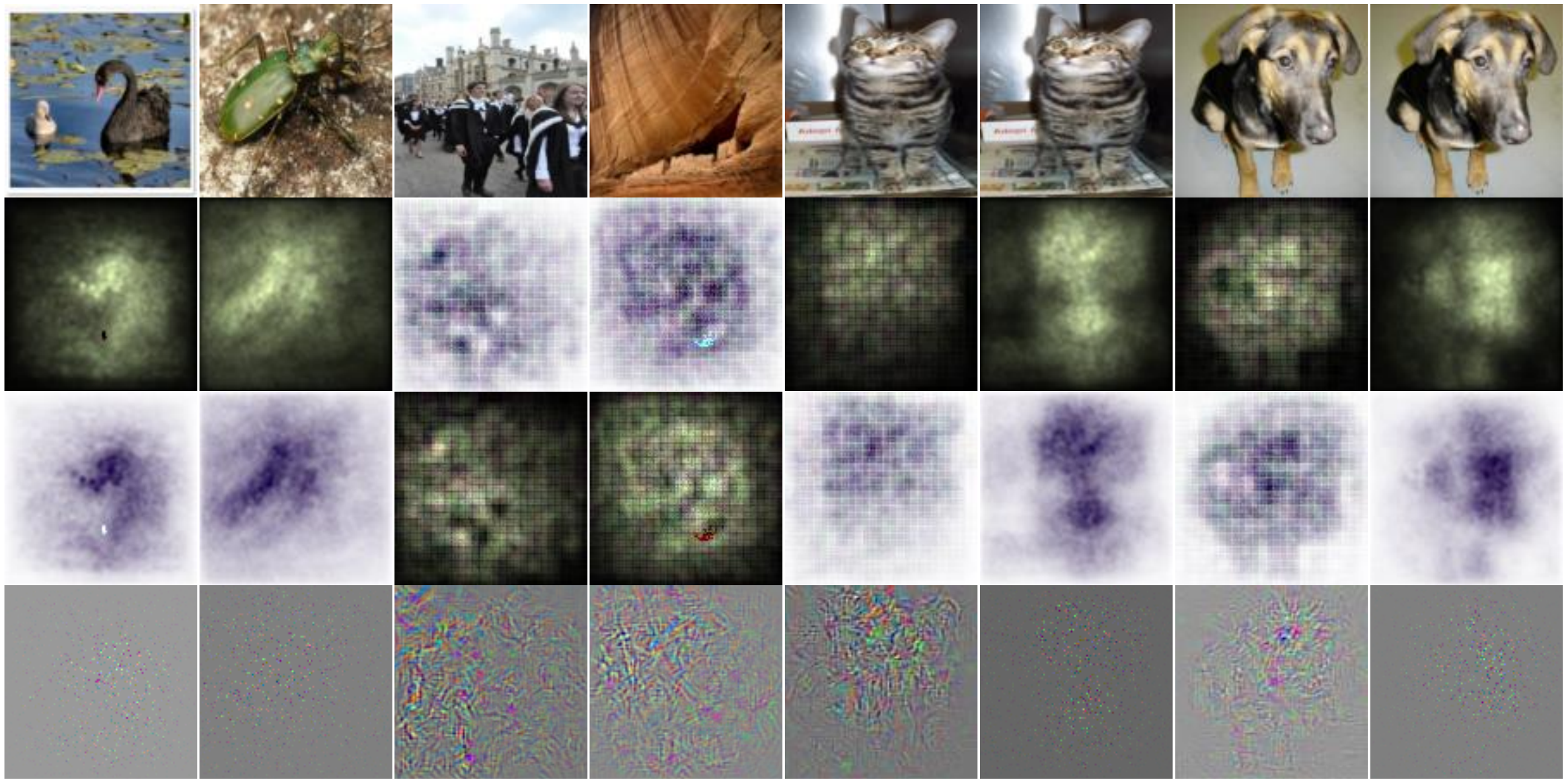}
	\label{visualization}
	\caption{Visualization of excitation activation maps. The presented images showcase the results of our proposed positive and negative excitation (PANE) method on a range of samples from ImageNet(ILSVRC) and Dogs. vs. Cats datasets. The first row displays the original images, while the second and third rows show the positive and negative excitation activation maps, respectively. The fourth row presents the visualization of the positive excitation plus the negative excitation, namely the addition map. Notably, the first four columns correspond to samples from ImageNet(ILSVRC), while the last four columns depict samples from the Dogs vs. Cats dataset. The activation maps generated for VGG16 are shown in columns 1, 2, 6, and 8, while the activation maps generated for AlexNet are displayed in columns 3, 4, 5, and 7.}
\end{figure*}

\subsubsection{Visualization}
The question arises: what exactly does negative excitation mean, and what is the relationship between positive and negative excitation? To shed light on this matter, we first present a sample visualization of the positive and negative excitation activation maps, as well as the addition map between them, which visualizes the addition of the positive excitation and the negative excitation. These visualizations are depicted in Figure 5.

\begin{table*}
	\centering
	\caption{Removing salient pixels for AlexNet and VGG-16 on Dog vs. Cats datasets. The method marked as Best is the best experimental result in Section 5.A. POS+NEG is the addition of positive and negative saliency maps.}\label{111}
	\begin{tabular}{lcccccccccccc}
		\hline
		datasets      & Networks & Method  & 0.1\%  & 0.2\%  & 0.3\%  & 0.4\%  & 0.5\%  & 0.6\%  & 0.7\%  & 0.8\%  & 0.9\%  & 1\%    \\
		\hline
		\multirow{4}{*}{Dogs vs. Cats} & \multirow{2}{*}{AlexNet}  & Best    & -.0330 & -.0503 & -.0657 & -.0797 & -.0898 & -.0983 & -.1064 & -.1146 & -.1224 & -.127  \\
		&  & POS+NEG & -.1281 & -.2116 & -.2677 & -.3091 & -.3445 & -.3722 & -.3959 & -.4204 & -.4339 & -.447  \\
		\cline{3-13}
		& \multirow{2}{*}{VGG-16}   & Best    & -.016  & -.0323 & -.0427 & -.0525 & -.0619 & -.0716 & -.0836 & -.0898 & -.0942 & -.1012 \\
		&   & POS+NEG & -.0119 & -.018  & -.0252 & -.0357 & -.0405 & -.0483 & -.0572 & -.0631 & -.0684 & -.0742 \\
		\hline
		\multirow{4}{*}{ImageNet1k}    & \multirow{2}{*}{AlexNet}  & Best    & -.0399 & -.0997 & -.1683 & -.2319 & -.2862 & -.3329 & -.3795 & -.418  & -.4594 & -.4959 \\
		&  & POS+NEG & -.0842 & -.204  & -.3237 & -.4276 & -.519  & -.5845 & -.6422 & -.6892 & -.7234 & -.7559 \\
		\cline{3-13}
		& \multirow{2}{*}{VGG-16}   & Best    & -.0573 & -.1251 & -.1916 & -.254  & -.3171 & -.374  & -.4261 & -.4699 & -.5073 & -.5412 \\
		&  & POS+NEG & -.051  & -.1141 & -.1618 & -.208  & -.2523 & -.2977 & -.3352 & -.3652 & -.3955 & -.4207 \\
		\hline
	\end{tabular}
\end{table*}

The visualization results of the positive and negative excitation activation maps are nearly complementary, which is consistent with intuition. However, when rechecking the specific values of the positive and negative excitation activation maps, it can be found that those coefficients with greater positive values correspond to the lower negative values. This means that the positive and negative excitation have opposite distributions. In other words, positive excitation and the inverse negative excitation are somehow equivalent. This derives a counter-intuitive result which is one pixel has both positive and negative excitation for the predicted outcome. This also makes the experimental results in the Section 5.A and Section 5.B obtained using the negative map the same as the positive map.
At the same time, it can be observed that, in some cases, the regions of the positive excitation and negative excitation logos feel like they are reversed, e.g. columns 3 and 4. This condition can be eliminated easily by simply checking the value, as one map only has positive or negative numbers. However, it is still an interesting phenomenon that may relate to the specific process of feature value transformation in networks. This is why we show the original results.

On the other side, as referred previously, it seems that if one region has a greater positive excitation value, it will also have a greater negative value. However, the addition maps reveal that the saliency maps for positive excitation and the opposite of negative excitation are fundamentally different.
This means that positive excitation and inverse negative excitation are somehow equivalent but not the same.

Based on these observations, we speculate that discrete pixel points may have a greater impact on the model's results compared to continuous regions. We also hypothesize that when the positive excitation of a pixel is greater than the negative excitation, that pixel will have a positive effect on the final result. To further verify our speculations, we experimentally verified the addition map, which refers to $\text{POS}+\text{NEG}$.

\subsubsection{Positive Excitation Greater Region}
When considering the saliency map composed of points with the addition map, removing salient pixels with the same settings as in Section 5.1 would intuitively lead to a more significant reduction in the model's prediction probability. To compare the best experimental results from Section 5.1 with $\text{POS}+\text{NEG}$, we extracted the relevant data and presented it in Table 1.

The experimental results demonstrate that $\text{POS}+\text{NEG}$ performs exceptionally well. In the low perturbation interval, such as a salient pixel removing ratio of 0.1\%, $\text{POS}+\text{NEG}$ achieves better performance compared with the best result from previous experiments on AlexNet. This substantial advantage is consistently maintained in high perturbed intervals, such as a salient pixel-removing ratio of 1\%. And, $\text{POS}+\text{NEG}$ is also very close to the best performance on VGG. It is worth to be noticed that the value of the best experimental result in Section 5.A is all from Guided BP. And, Guided BP has a very bad performance on the experiment in Section 5.B, which refers to the Guided BP might point to the sensitivity. $\text{POS}+\text{NEG}$, which points to the saliency, has a similar or better performance than Guided BP. Joint observing Table 1 and Fig. 3, it can be observed that $\text{POS}+\text{NEG}$ completely overperforms other methods, including PANE-POS. Based on these results, we can reasonably infer that $\text{POS}+\text{NEG}$ is more effective at identifying the most salient points for the network.

\begin{table*}
	\centering
	\small
	\caption{Removing negative salient pixels from POS+NEG for AlexNet and VGG-16 on Dog vs. Cats datasets. }\label{222}
	\begin{tabular}{lccccccccccc}
		\hline
		Networks & 0.1\%  & 0.2\% & 0.3\%  & 0.4\%  & 0.5\%  & 0.6\%  & 0.7\%  & 0.8\%  & 0.9\% & 1\%    \\
		\hline
		AlexNet  & .0197  & .0206 & .0205  & .0171  & .0161  & .015   & .0136  & .0124  & .0109 & .0097  \\
		VGG-16   & -.0097 & -.02  & -.0278 & -.0376 & -.0491 & -.0551 & -.0644 & -.0699 & -.076 & -.0826 \\
		\hline
	\end{tabular}
\end{table*}

\subsubsection{Guiding adversarial attack}
We believe that a high-quality saliency map should be able to effectively guide downstream tasks, such as generating adversarial samples. To this end, we attempted to utilize the saliency map to guide the adversarial attack method in generating inconspicuous perturbations. First, the adversarial attack method was utilized to generate an adversarial perturbation for a given image. Then, the saliency map was used to retain the perturbations corresponding to the higher saliency regions in the original images. Finally, the retained perturbations were added to the original image and fed into the network. If a saliency map is an effective guide to the adversarial attack method, then the retained perturbations should disable the network as much as possible. We applied I-FGSM \cite{IFGSM}, one of the most classic adversarial attack methods, to generate adversarial perturbations, with AlexNet and VGG-16 as the victim networks and the Dogs vs. Cats dataset. The $L_{\infty}$ of the perturbation was set to 50, the step size to 7, and the number of iterations to 10, ensuring that every adversarial perturbation resulted in a successful attack. We compared the results of PANE-POS and POS+NEG with the baselines of Grad-CAM and Grad-CAM++, with the percentage of a reserved area set at 0.5\%-3\%. The experimental results are presented in Figure 6.

\begin{figure}
	\centering
	\includegraphics[width=8.7cm]{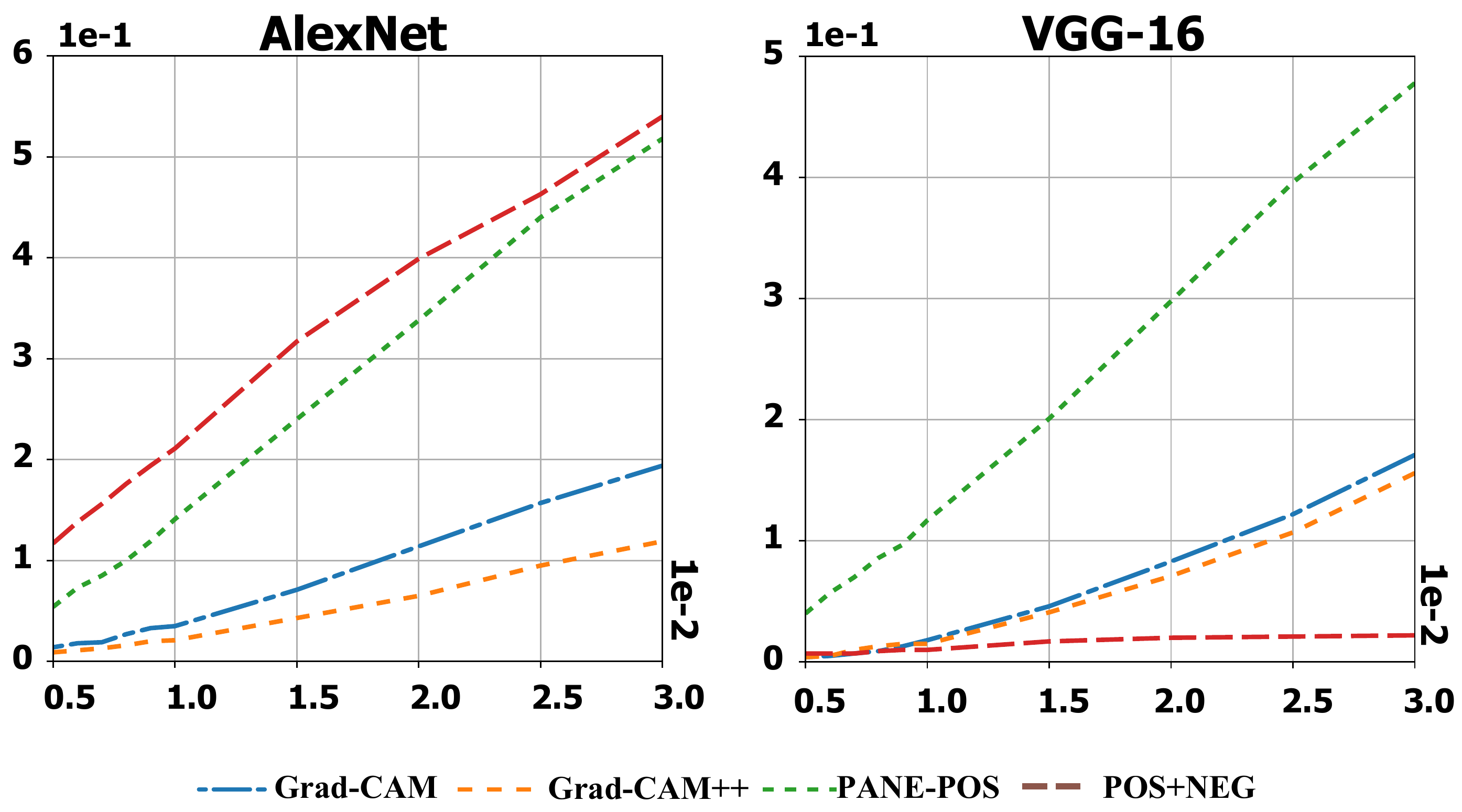}
	\caption{Experimental results on guiding adversarial attack. The network corresponding to the results of that experiment is indicated above each subplot.}
\end{figure}

The experimental results indicate that PANE-POS has a significant advantage in guiding sample generation. On both AlexNet and VGG-16, PANE-POS achieves several times the attack success rate of Grad-CAM and Grad-CAM++. However, a curious phenomenon can be observed when examining the results of the guidance from POS+NEG. While POS+NEG achieved better results than PANE-POS on AlexNet, it performed poorly on VGG-16. When combined with the difference map in Figure 5, we suspect that this may be due to the discontinuity of POS+NEG, which caused the failure of the adversarial perturbation retained on VGG-16. This may also suggest that the correlation between adversarial attacks and salient pixels requires further exploration.

\subsubsection{Negative Excitation Greater Region}
A natural question raised is what is the characteristic of the negative value in the addition map?
Intuitively, removing the most negative salient pixels of the saliency map can verify the characteristic and this action will increase the confidence of the model's output for the corresponding class. We focus on studying POS+NEG under the binary classification task, specifically on the Dogs vs. Cats dataset, to eliminate the interference of interclass correlation. The experiment is conducted under the same settings as in Section 5.1, and the results are presented in Table 2.

Examining the data for AlexNet, we observe that removing the salient pixels in POS+NEG can indeed improve the prediction probability of the target class. However, it is noteworthy that the probability increase is highest at a removal rate of 0.2\%. As the removal rate increases, the value of the probability increase becomes smaller and smaller. This may be attributed to the effect that as more pixels are removed, the negative excitation of the removal cannot eliminate the influence from the removed pixels, i.e., the black blanks in the image. However, when examining the data for VGG-16, a different situation seems to occur.
The prediction probability of VGG-16 decreases even more than PANE-POS after removing the negative salient pixels in POS+NEG. Even if we remove only the most negative salient pixel, the prediction probability of VGG-16 still decreases. Comparing the difference maps of AlexNet and VGG-16 in Figure 5, we speculate that the phenomenon may be because the saliency pixels of POS+NEG for VGG-16 mainly focus on the target object. This makes removing that pixel have a greater impact on the surrounding pixels. This also raises another question to ponder, as there may be higher-order correlations between positive and negative excitation when the network becomes deeper.

\subsubsection{Logit Changes}

\begin{table*}
	\centering
	
	\caption{Logit changes with reducing salient pixels. The first row represents the perturbation region ratio. The second, third, sixth, and seventh rows represent the accumulation of the logit value of the perturbed image minus the original image's logit value. The fourth, fifth, eighth, and ninth rows represent the ratio of the number of samples with decreased logit values to the total number of samples.}\label{logitchangessalient}
	\begin{tabular}{lccccccccccc}
		\hline
		datasets      & Method  & 0.01\% & 0.02\%  & 0.03\%  & 0.04\%  & 0.05\%  & 0.06\%  & 0.07\%    & 0.08\%  & 0.09\%   & 0.1\%   \\
		\hline
		\multirow{4}{*}{Dogs vs. Cats} & POS     & 0.055  & 0.11    & 0.0806  & 0.177   & 0.152   & 0.206   & 0.219     & 0.288   & 0.312    & 0.349   \\
		& POS+NEG & -4.048 & -8.528  & -12.652 & -16.491 & -19.232 & -22.751 & -26.146& -29.474 & -31.938  & -35.157 \\
		\cline{2-12}
		& POS     & 49.7\%  & 49.7\%   & 47.4\%   & 48.3\%   & 47.9\%   & 47.6\%   & 48.1\%     & 47.3\%   & 47.6\%    & 48.2\%   \\
		& POS+NEG & 100\%    & 100\%     & 100\%     & 100\%     & 100\%     & 100\%     & 100\%       & 100\%     & 100\%      & 100\%     \\
		\hline
		\multirow{4}{*}{ImageNet1k}  & POS     & -0.015 & -0.030  & -0.028  & -0.114  & -0.146  & -0.116  & -0.135    & -0.068  & -0.159   & -0.126  \\
		& POS+NEG & -6.233 & -13.259 & -19.666 & -25.663 & -30.024 & -35.479 & -40.920   & -46.041 & -49.811 & -54.816 \\
		\cline{2-12}
		& POS     & 42.2\%  & 41.5\%   & 44.1\%   & 43.8\%   & 45.6\%   & 45.6\%   & 46.1\%     & 44.8\%   & 46.2\%    & 45.7\%   \\
		& POS+NEG & 99.8\%  & 100\%     & 100\%     & 100\%     & 100\%     & 100\%     & 100\%       & 100\%     & 100\%      & 100\%     \\
		\hline
	\end{tabular}
\end{table*}

\begin{table*}
	\centering
	
	\caption{Logit Changes with reducing minor pixels. The first row represents the perturbation region ratio. The second, third, sixth, and seventh rows represent the accumulation of the logit value of the perturbed image minus the original image's logit value. The fourth, fifth, eighth, and ninth rows represent the ratio of the number of samples with increased logit values to the total number of samples.}\label{logitchangesminor}
	\begin{tabular}{lccccccccccc}
		\hline
		datasets      & Method  & 0.01\% & 0.02\% & 0.03\% & 0.04\% & 0.05\% & 0.06\% & 0.07\% & 0.08\% & 0.09\% & 0.1\%  \\
		\hline
		\multirow{4}{*}{Dogs vs. Cats} & POS     & 0.003  & -0.006 & 0.002  & -0.007 & -0.007 & 0.001  & 0.003  & -0.001 & -0.002 & -0.001 \\
		& POS+NEG & 4.051  & 8.799  & 13.095 & 17.033 & 19.899 & 23.493 & 27.019 & 30.402 & 32.858 & 36.093 \\
		\cline{2-12}
		& POS     & 36.3\%  & 38.3\%  & 42.6\%  & 38.7\%  & 40.7\%  & 43.1\%  & 43.8\%  & 42.0\%   & 45.2\%  & 43.3\%  \\
		& POS+NEG & 99.9\%  & 100\%    & 100\%    & 100\%    & 100\%    & 100\%    & 100\%    & 100\%    & 100\%    & 100\%    \\
		\hline
		\multirow{4}{*}{ImageNet1k} & POS     & -0.038 & -0.001 & 0.080  & 0.057  & 0.071  & 0.189  & 0.145  & 0.191  & 0.221  & 0.202  \\
		& POS+NEG & 6.256  & 13.107 & 19.476 & 25.538 & 29.828 & 35.374 & 40.670 & 45.824 & 49.506 & 54.329 \\
		\cline{2-12}
		& POS     & 39.1\%  & 40.7\%  & 42.8\%  & 43.1\%  & 44.0\%   & 43.8\%  & 43.5\%  & 43.9\%  & 44.5\%  & 42.4\%  \\
		& POS+NEG & 99.9\%  & 100\%    & 100\%    & 100\%    & 100\%    & 100\%    & 100\%    & 100\%    & 100\%    & 100\%    \\
		\hline
	\end{tabular}
\end{table*}

Considering PANE is proposed directly target to the output of the network, which is usually the logit value, it is necessary to verify whether excitations can correctly indicate the saliency for those logit, but not the classification probabilities. To narrow the range of perturbation, this verification reduces the perturbation region and only reduces 1 to the value of each selected pixel, which ranges from 0 to 255. Because the depth of AlexNet is relatively not great, the fixed bias of AlexNet will not be very great. Thereby, we select AlexNet as the candidate network to verify the logit changes in both reduced positive salient pixels and reduced negative salient pixels. Following the intuition, if positive salient pixels are reduced, the logit value of the corresponding class will drop. And, if negative salient pixels are reduced, the logit value of the corresponding class will increase. Both the total of logit changes and the number of changes are the same as the idea changes are counted. Experimental results for these two reductions are shown in TABLE 3 and TABLE 4.

It can be observed that POS+NEG completely meets expectations. In TABLE 3, no matter Dogs vs. Cats or ImageNet1k, all changes of logit values from POS+NEG are negative values, which means logit value drops. Except for two images in ImageNet1k under 0.01\%, the logit value of all other samples drops. While only about half of the samples from PANE-POS gain the expected result. In TABLE 4, although some logit value changes of PANE-POS are positive numbers that correspond to the expectations, the number of samples whose logit value increases is not satisfactory. While, POS+NEG still gains a superb performance, as all results are positive value and almost all samples gain increasing logit value. With these experimental results, it can be inferred that PANE, especially POS+NEG, indeed can effectively reflect the saliency for the logit values to a certain extent.

\section{Conclusion and Future work}
This paper introduces a novel concept called positive and negative excitation (PANE), which is capable of decomposing each layer into a positive and negative excitation in local explanation. A double-chain backpropagation process is proposed to obtain a combination of excitation coefficients between any two layers, resulting in the generation of positive and negative excitation maps. Experimental results demonstrate that the proposed PANE method significantly outperforms all five baselines. Moreover, in further correlation experiments, our proposed method is found to be more effective in guiding downstream tasks such as adversarial attacks.
In addition to the promising results, some interesting phenomena were also discovered. The first phenomenon is that each pixel serves as both the positive and negative excitation signal for the final result, providing a new perspective on the interpretation of neural networks. The second phenomenon is the possibility of a higher-order relationship between positive and negative excitation maps, which warrants further investigation to better understand the impact of model inputs on outputs.

In future advancements, we plan to enhance PANE from two perspectives to address its current limitations. Firstly, we aim to simplify the implementation process of PANE while maintaining its effectiveness, eliminating the need for extensive modifications to existing deep learning frameworks. Additionally, we intend to extend PANE beyond the real number field by incorporating group theory and functional analysis. This expansion will enhance PANE's adaptability and accuracy, allowing it to overcome the constraints imposed by the characteristic limitations and the complete real number space.

Furthermore, we plan to delve deeper into this correlation to gain a more comprehensive understanding of the underlying mechanisms of neural networks. The proposed PANE method has shown great potential in improving the interpretability of neural networks and guiding downstream tasks, and we believe that further exploration of this concept will pave the way for more advanced and reliable machine learning models.

\section*{Acknowledgments}
This work has been supported by the National Key Research and Development Program of China under grants 2020YFB1804604, and the University-Industry Collaborative Education Program under grants 220602842235333.

\bibliographystyle{abbrv}
\bibliography{nips}

\begin{IEEEbiography}[{\includegraphics[width=1in,height=1.25in, clip, keepaspectratio]{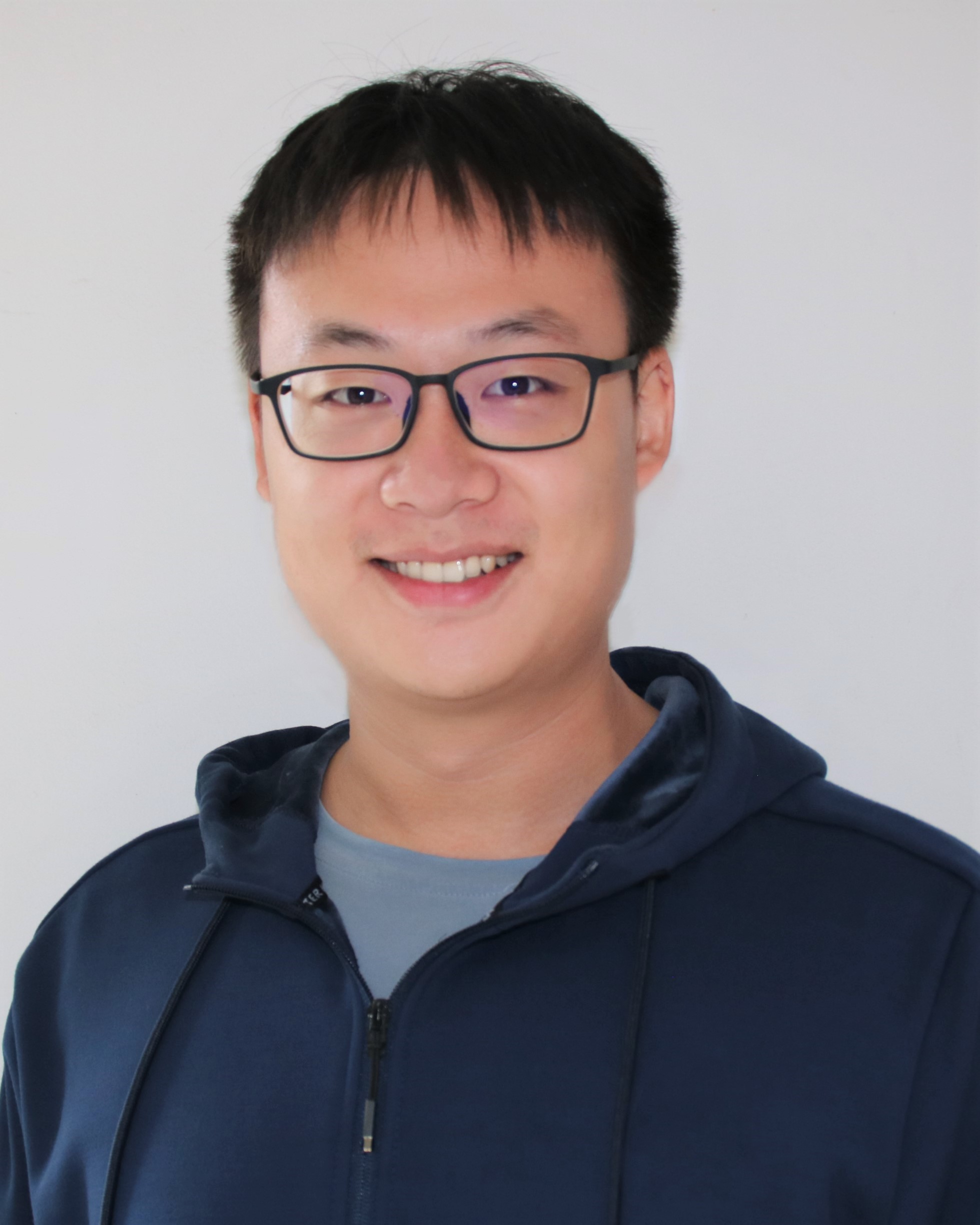}}]{Zijian Ying}received the B.S. degree from Nanjing University of Science and Technology in 2019. He is currently pursuing a Ph.D. degree at the School of Computer Science and Engineering, Nanjing University of Science and Technology, Nanjing, China. His research interests include machine learning, interpretable deep learning, cyber security, data mining, and crowdsourcing.
\end{IEEEbiography}

\begin{IEEEbiography}[{\includegraphics[width=1in,height=1.25in,clip,keepaspectratio]{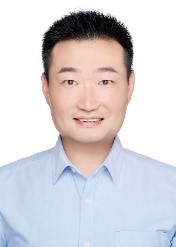}}]{Qianmu Li}
	, professor and doctoral supervisor; foreign academician of the Russian Academy of Natural Sciences; member of Academic Committee of Nanjing University of technology, director of information construction and Management Department of Nanjing University of technology; Vice chairman of Jiangsu Science and Technology Association. Academic member of Cyber Engineering Laboratory of State Grid, deputy director of intelligent education special committee of National Computer Basic Teaching and Research Association, chief expert of e-government platform in Jiangsu Province, vice chairman of Jiangsu Digital Government Standardization Technical Committee, President of Nanjing Computer Society, vice president of Jiangsu Cyber Engineering Society, executive director of Jiangsu Computer Society, Secretary General of Jiangsu Internet Finance Association. He was selected as the first network and security outstanding talent of China communication society, the young and middle-aged leading talent of Jiangsu Province. He has won more than ten first and second prizes, including the science and technology progress award of the Ministry of education, the science and technology award of Jiangsu Province, the teaching achievement award of Jiangsu Province, the outstanding achievement award of scientific research in universities of the Ministry of education, and the best paper awards such as ISKE, AAAI and ICCC.
\end{IEEEbiography}

\begin{IEEEbiography}[{\includegraphics[width=1in,height=1.25in, clip, keepaspectratio]{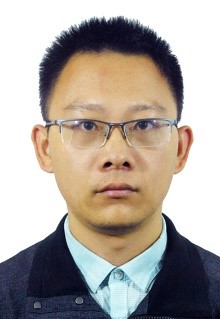}}]{Zhichao Lian} received the bachelor's and master's degrees in computer science from Jilin University, Changchun, China, in 2005 and 2008, respectively, and the Ph.D. degree from Nanyang Technological University in 2013. From 2012 to 2014, he was a Post-Doctoral Associate with the Department of Statistics, Yale University. He is currently an Associate Professor with the School of Cyber Science and Engineering, Nanjing University of Science and Technology, China. His research areas include image processing, pattern recognition, and artificial intelligence.
\end{IEEEbiography}

\begin{IEEEbiography}[{\includegraphics[width=1in,height=1.25in, clip, keepaspectratio]{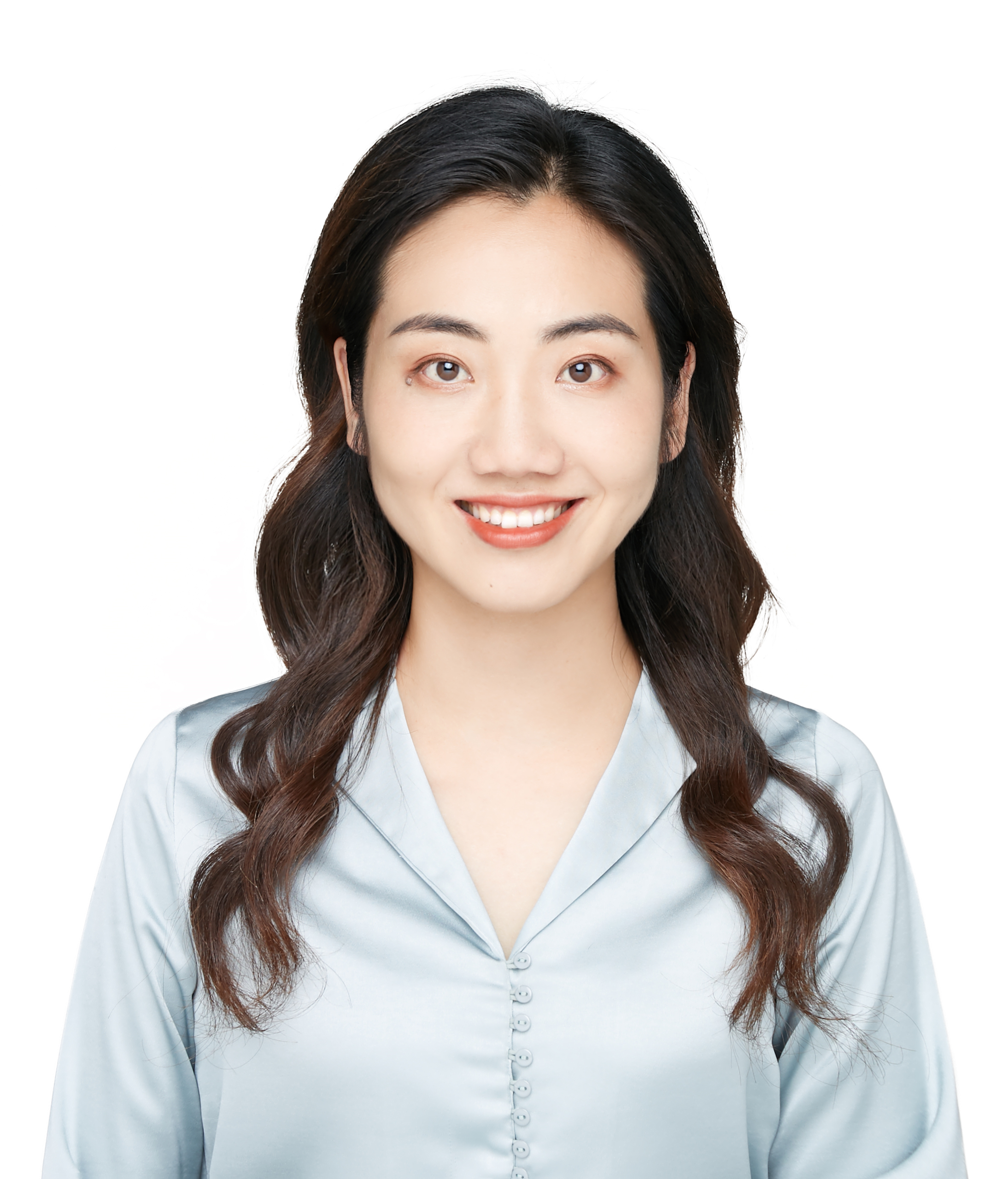}}]{Jun Hou}, associate research fellow; director of Teaching and Research Office, Department of Social Sciences, Nanjing Vocational University of Industry Technology; specially-appointed researcher at Zijin College,  Nanjing University of Science and Technology; deputy secretary general of the Jiangsu Provincial Association for International Science and Technology Development;  member of Jiangsu Qinglan Project; the leader of a major philosophy and social science project of universities in Jiangsu province; main drafter of two local standards (DB 32/T 4274, DB32/T 3875) in the field of digital economy in Jiangsu Province. She has won the third prize of Jiangsu science and technology award and the second prize of science and technology award of China communication society. Her research interests is information society and research results focus on regional digital innovation management.
\end{IEEEbiography}

\begin{IEEEbiography}[{\includegraphics[width=1in,height=1.25in, clip, keepaspectratio]{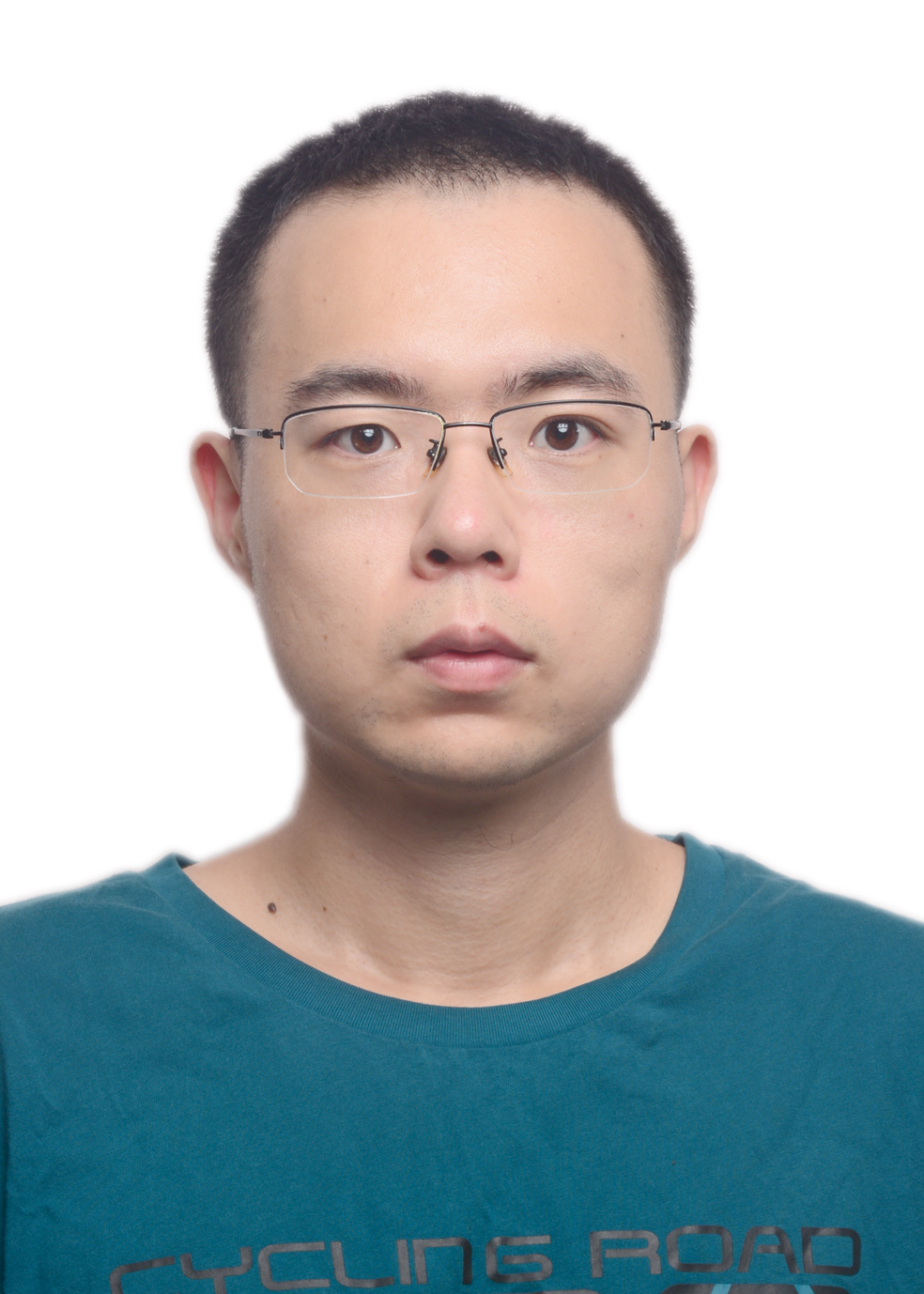}}]{Tong Lin} received the B.S. degree from Peking University in 2016. He is currently pursuing the Master degree at the School of Computer Science and Engineering, Nanjing University of Science and Technology, Nanjing, China. His research interests include machine learning, graph representation learning and disentangled representation learning.
\end{IEEEbiography}

\begin{IEEEbiography}[{\includegraphics[width=1in,height=1.25in, clip, keepaspectratio]{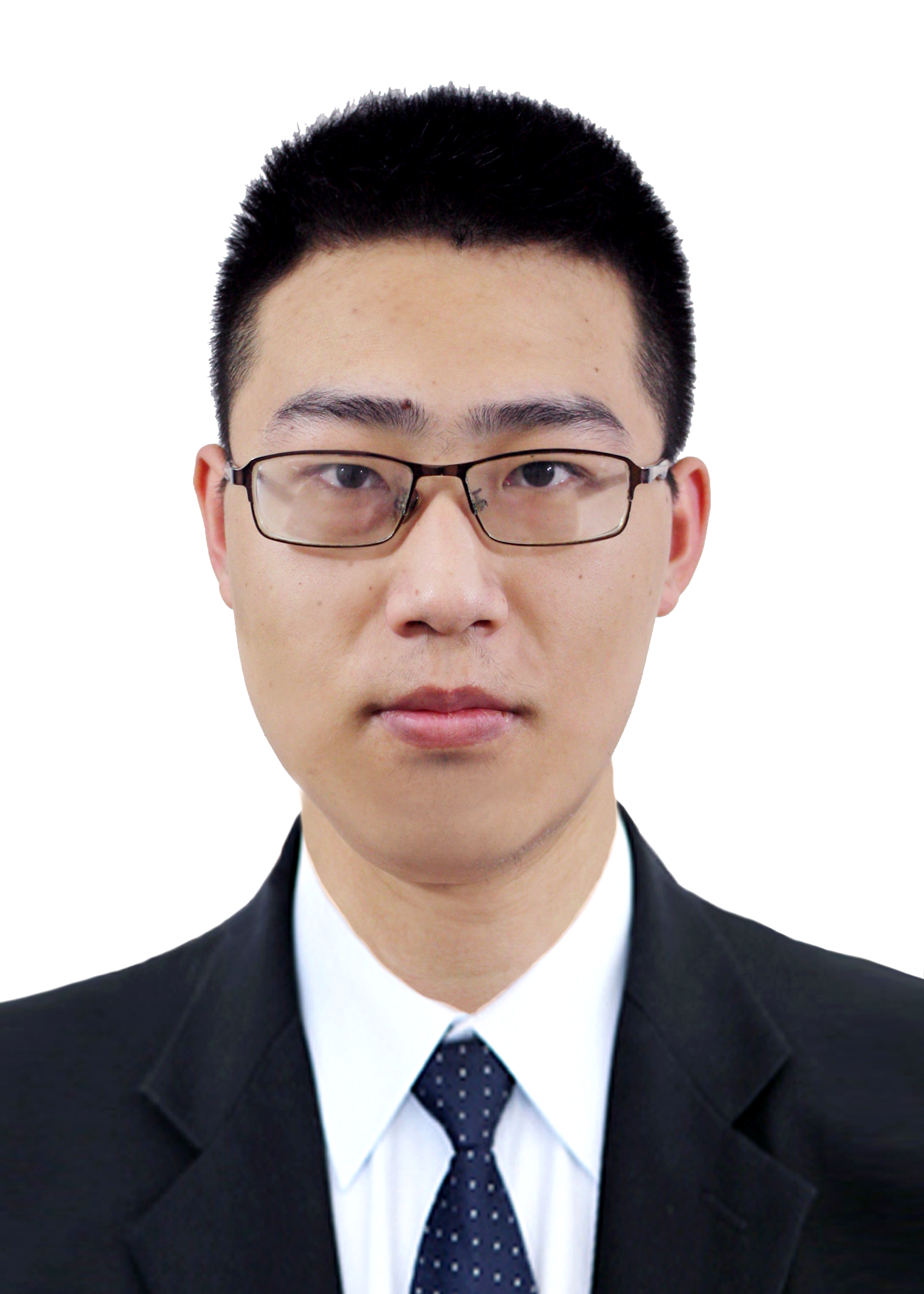}}]{Tao Wang}  received the B.S. degree from Nanjing University of Science and Technology in 2022.He is currently pursuing the Master degree at the School of Software Engineering,Nanjing University of Science and Technology,Nanjing,China.His research interests include machine learning, adversarial attack and transferability.
\end{IEEEbiography}

\end{document}